\documentclass{bmvc2k}

\usepackage{multirow}
\usepackage{xspace}
\usepackage{booktabs}
\usepackage{amsmath}
\usepackage{amssymb}
\DeclareGraphicsExtensions{.pdf}

\newcommand{\mb}[1]{\underline{#1}}

\title{Concise Radiometric Calibration Using The Power of Ranking}

\addauthor{Han Gong$^*$}{h.gong@uea.ac.uk}{1}
\addauthor{Graham D. Finlayson$^*$}{g.finlayson@uea.ac.uk}{1}
\addauthor{Maryam M. Darrodi}{}{1}

\addinstitution{
 School of Computing Sciences\\
 University of East Anglia\\
 Norwich, UK\\
 $^*$ denotes co-first author
}

\runninghead{Gong, Finlayson, Darrodi}{Concise Radiometric Calibration by Ranking}

\def\eg{\emph{e.g}\bmvaOneDot}
\def\ie{\emph{i.e}\bmvaOneDot}

\def\etal{\emph{et al}\bmvaOneDot}

\begin{document}

\maketitle

\begin{abstract}
Compared with raw images, the more common JPEG images are less useful for machine vision algorithms and professional photographers because JPEG-sRGB does not preserve a linear relation between pixel values and the light measured from the scene. A camera is said to be radiometrically calibrated if there is a computational model which can predict how the raw linear sensor image is mapped to the corresponding rendered image (\eg JPEGs) and vice versa. This paper begins with the observation that the rank order of pixel values are mostly preserved post colour correction. We show that this observation is the key to solving for the whole camera pipeline (colour correction, tone and gamut mapping). Our rank-based calibration method is simpler than the prior art and so is parametrised by fewer variables which, concomitantly, can be solved for using less calibration data. Another advantage is that we can derive the camera pipeline from a single pair of raw-JPEG images. Experiments demonstrate that our method delivers state-of-the-art results (especially for the most interesting case of JPEG to raw).
\end{abstract}

\section{Introduction}
Many computer vision algorithms (\eg photometric stereo~\cite{Petrov.shape.CRA}, photometric invariants~\cite{GeversFS04}, shadow removal~\cite{shadow16}, and colour constancy~\cite{Koberooni2}) assume that the captured RGBs in images are linearly related to the actual scene radiance. However, the imaging pipeline in a digital camera is necessarily non-linear in order to produce perceptually-pleasing photos rather than its physically-meaningful counterparts. In this paper, we present a new rank-based radiometric calibration method which solves for the bi-directional mappings between the camera's RAW responses and the rendered RGBs produced by digital camera.

There is prior art in this field which models the pipeline with a large number of parameters (up to several thousand~\cite{chak}) which both means a large corpus of data is required to uncover the pipeline and that there is at least tacitly the premise that the underlying pipeline is quite complex. The key insight in our approach is that post-colour correction (a $3 \times 3$ matrix correction) the linear corrected raw RGBs are to the greatest extent in the same rank order as the final rendered RGBs. Building on this insight, we develop a simple rank-based radiometric calibration model that ``solves for'' the camera pipeline with many fewer parameters and concomitantly needs much less training data.

\begin{figure}[htb]
\begin{center}
   \includegraphics[width=\linewidth]{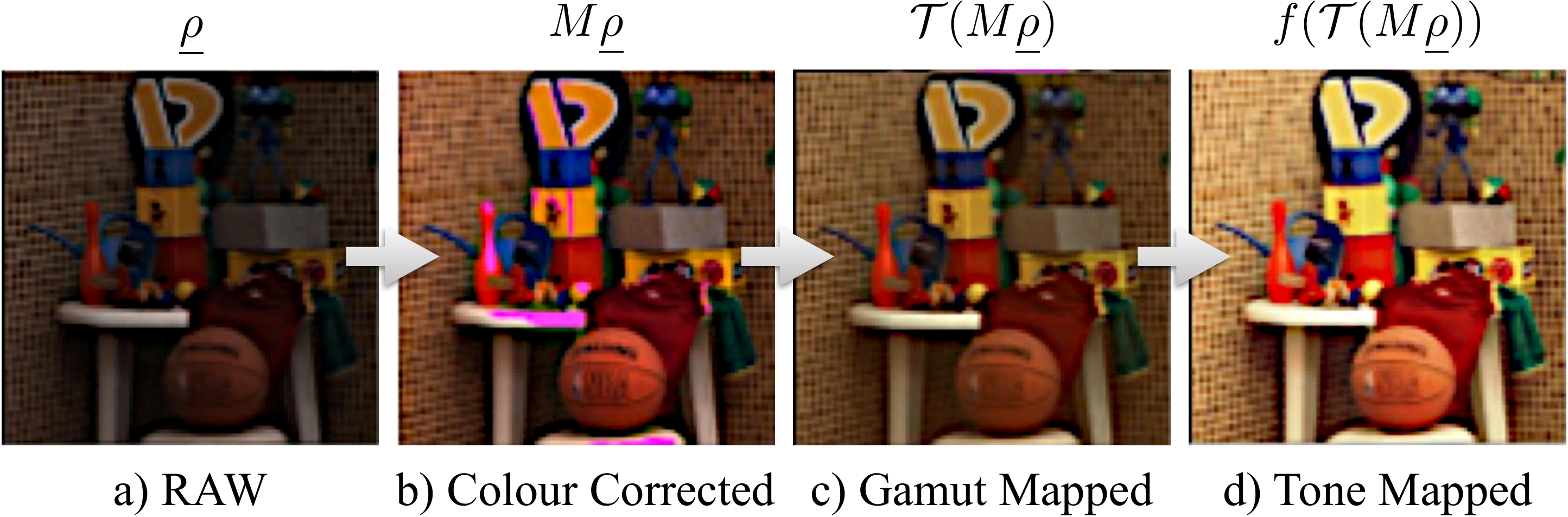}
\end{center}
   \caption{a) a RAW input image is colour corrected to give image b). Non-displayable colours are highlighted in purple pseudo colour. Gamut mapping, in step c), brings colours within gamut. Finally, in d), a tone mapping step results in the final rendered image.}
\label{fig:RC-PP}
\end{figure}

In Fig.~\ref{fig:RC-PP}, we illustrate a conventional image reproduction pipeline that holds for many cameras~\cite{BrownPami12}. An exemplar raw image, Fig.~\ref{fig:RC-PP}a, is mapped by a $3 \times 3$ colour correction matrix to give the image shown in Fig.~\ref{fig:RC-PP}b. The colour correction matrix implements several processing steps (\eg illumination correction~\cite{PIPELINE,ChakBMVC}, display RGB mapping \cite{srgb}, and colour preference adjustments~\cite{PIPELINE}). It is well-known that a display device cannot display all captured image colours that some RGBs will fall outside the RGB cube after mapping (\eg the pixels marked in purple in Fig.~\ref{fig:RC-PP}b). We therefore need gamut mapping, \eg \cite{BrownPami12,chak,ZollikerGM}, to bring the colours back inside the cube as shown in Fig.~\ref{fig:RC-PP}c. Finally, the gamut mapped image is tone mapped to arrive at the final rendered output \cite{PIPELINE,ChakBMVC,BrownPami12} shown in Fig.~\ref{fig:RC-PP}d. Tone mapping accounts for the display non-linearity~\cite{srgb}, dynamic range compression and some aspects of preference~\cite{DRC}.

The colour processing pipeline -- for cameras, in general, can be written as Eqn.~\ref{eq:base_model}.
\begin{equation}
\begin{array}{cccccc}
    \mb{P} =\ & \underbrace{ {f}(\Gamma (M\mb{\rho})) } & =& \underbrace{ \Gamma({f}(M\mb{\rho}))} & \approx & \underbrace{\text{LUT}(\mb{\rho})} \\
& (\text{1a}) & & (\text{1b}) & & (\text{1c})
\end{array}
\label{eq:base_model}
\end{equation}
Here and throughout this paper, $\mb{\rho}$ denotes a camera RAW and $\mb{P}$ refers to its rendered RGB counterpart. Respectively, the $3 \times 3$ correction matrix, gamut mapping and tone mapping are denoted by the matrix $M$ and the functions $\Gamma()$ and ${f}()$. The function $f()$ can implement a single or three per-channel tone curves. Because gamut mapping only implements a small change in comparison with colour and tone mapping steps, the order of gamut mapping and tone mapping may be switched (Eqn.~\ref{eq:base_model}b \& c), a property that we exploit in this paper. Equally, we can also roll all three processing steps into one and directly solve for a 3D LUT (Look-Up-Table) that maps RAW to rendered counterparts. This LUT function is denoted $\text{LUT}()$~\cite{BrownECCVLattice12} in Eqn.~\ref{eq:base_model}c. Readers may refer to the top row of Fig.~\ref{fig:RC-PP} to link each mathematical function to our example processed image.

In radiometric calibration, given a set of $\mb{\rho}$ and $\mb{P}$, one seeks to solve for the parametrised pipeline parts (\eg $M$, $\Gamma()$, $f()$ and $\text{LUT}()$). A disadvantage of the current best performing methods is that a great deal of data may be required to fit their assumed models. In Eqns.~\ref{eq:base_model}a and \ref{eq:base_model}b, the gamut mapping step could be modelled by 1000s of Radial Basis functions~\cite{BrownPami12,BrownECCVLattice12,chak} and in Eqn.~\ref{eq:base_model}c, the deployed LUT could also have several thousand control points. 

Our proposed method begins with the simple observation~\cite{finlayson2016rank} that, assuming the gamut mapping step makes small changes to image colours, we expect the rank ordering of the rendered $\mb{P}$ to be the same as $\mb{\rho}$ multiplied by the correction matrix $M$ (because a tone curves are always monotonically increasing). Suppose that two rendered (JPEG) responses -- in the 1$^\text{st}$ red colour channel -- are denoted $P^a_1$ and $P_1^b$; and that $P^a_1>P^b_1$. The rank order of two corresponding raw red channel measurements post colour correction is written as $M_1\mb{\rho}^a>M_1\mb{\rho}^b$ (where $M_1$ denotes the first row of $M$ and $\mb{\rho}^a$ and $\mb{\rho}^b$ are a pair of raw RGBs). Rewriting: this implies that $M_1(\mb{\rho}^a-\mb{\rho}^b)>0$ which mathematically defines a half-space constraint. If we visualise the row vector $M_1$ as a point in 3-space then this inequality -- which we call a ranking constraint -- forces the point to be located in one half of 3-space but not the other. Because we have multiple pixels, each pair of pixels (2 raw and 2 JPEG RGBs) generates a half space constraint and intersecting all these constraints delimits the region in which $M_1$ must lie. Our experiments demonstrates that a small numbers of patches suffices to estimate $M$ accurately. Once we have $M$ we then find the best rank preserving tone curves $f()$. At this stage, only using $M$ and $f()$ we have a tolerable approximation of the pipeline. Indeed, we argue that our construction of $M$ and $f()$ also incorporates, to a first order, gamut mapping. Now we adopt (Eqn~\ref{eq:base_model}c) and find a 125-parameter LUT to ``mop up'' any remaining errors due to gamut mapping (higher order terms). 


\section{Related work}

Using the pipeline form Eqn~\ref{eq:base_model}b, Chakrabarti \etal~\cite{chak} first solve for $M$ and $f()$ (in a least-squares sense) in iteration and then solve directly for $\Gamma()$. In their approach, $f()$ is constrained to be a 7$^\text{th}$ order monotonic polynomial. They model $\Gamma()$ with the radial basis function (RBF) method of \cite{BrownPami12} where several thousands of RBFs are potentially used. A restriction of the above calibration is presented in \cite{ChakBMVC} where  the gamut mapping $\Gamma()$ is ignored. This less general model works tolerably well on many real pairs of raw and rendered images and this is a point we will return to later in this paper.
In either version (\cite{chak} or \cite{ChakBMVC}), the coupled nature of the minimization means that a global minimum is not guaranteed to be found. Thus, a random start search is incorporated -- multiple minimisations are carried out -- in order to find their best parameters. 
Kim \etal \cite{BrownPami12} solve for the pipeline in the form of Eqn.~\ref{eq:base_model}a and makes additional assumptions to decouple the optimization. They assume that images of the same scene are captured with respect to two or more exposures and their $\Gamma()$ is a multi-thousand set of RBFs. Regarding solving for $f()$, Debevec \etal~\cite{Debevec97} showed how relating corresponding pixels under known exposure differences suffices to solve for $f()$ (assuming there is no gamut mapping step). Importantly, in \cite{BrownPami12}, it was argued that for the set of desaturated pixels (\ie RAWs far from the RGB cube boundary) the gamut mapping step has little or no effect and can be ignored.  Relative to this assumption, $f()$ can be solved using the Debevec method. Given $f()$ then the colour correction matrix $M$ can be found (again using desaturated pixels). Though, for typical capture conditions, e.g. for most mobile phones,  multiple exposures are not available and so the requirement that multiple exposures are needed is a weakness in this method. Finally, in \cite{BrownPami12} a gamut mapping RBF network is ``trained''.
Of course, if a large number of radial basis functions are used to model gamut mapping (as proposed in \cite{BrownPami12} or \cite{chak}) then solving for $\Gamma()$ requires a large corpus of data. Further the application of gamut mapping is expensive and its parametrisation is large.

In~\cite{BrownECCVLattice12} it was shown that is possible to ignore the underlying structure of the colour processing pipeline and directly solve for the best 3D surjective function -- implemented as a LUT that maps the RAWs to rendered RGBs (Eqn.~\ref{eq:base_model}c).
Finally, in \cite{RadiometricEdge}, a method is presented for solving for $f()$ by examining the edge distribution in an image. This method has the advantage that the method works for a single image (no need for multiple exposures) but the disadvantage that the method is sensitive to processing steps such as image sharpening which is used extensively in mobile phone image processing.

\section{The rank-based method}

As the reader shall see, to make the rank-based method work we need to assume that the gamut mapping step $\Gamma()$ only makes small adjustments to colour. In fact our assumption is more nuanced. We assume that -- to a first order -- gamut mapping can mostly be implemented as an affine transform and that this affine transform can be folded into the colour correction  matrix $M$ and the  monotonically increasing tone mapping functions $f()$. 

\subsection{Gamut Mapping as an Affine Transform}

In Eqn.~\ref{eq:base_model}b, gamut mapping is applied when, after colour correction, colours are mapped outside the colour cube and become non-displayable. 
Let us use a Taylor expansion to model $\Gamma()$ around a point $\mb{a}$ inside the gamut:
\begin{equation}
\label{eq:gamma}
\Gamma(M\mb{\rho})\approx \Gamma(\mb{a})+J(\mb{a})(\mb{\rho}-\mb{a})
\end{equation}
where $J$ is the $3 \times 3$ Jacobian (matrix of derivatives of $\Gamma$). Not only does Eqn.~\ref{eq:gamma} show that, to a first approximation, gamut mapping is an affine transform it is also one of the gamut mapping algorithms proposed in \cite{ZollikerGM}. In particular, \cite{ZollikerGM} solves, with good results, for the best affine transform that maps image colours inside the gamut and which are, simultaneously, close to the non-gamut mapped colours:
\begin{equation}
\min_{T,\mb{o}} \Sigma_i ||TM\mb{\rho}_i+\mb{o}-M\mb{\rho}||^2\;\;s.t.\;\mb{0}\leq TM\mb{\rho}_i+\mb{o}\leq\mb{1}
\label{eq:gamut_min}
\end{equation}
In Eqn.~\ref{eq:gamut_min}, $T$ and $\mb{o}$ are respectively a $3 \times 3$ matrix and $3 \times 1$ offset vector defining the affine gamut mapping algorithm. The 3-vectors of 0s and 1s are denoted $\mb{0}$ and $\mb{1}.$ Eqn.~\ref{eq:gamut_min} is solved directly by Quadratic Programming~\cite{Gill81}. The gamut mapping shown in Fig.~\ref{fig:RC-PP}c is the result of solving Eqn.~\ref{eq:gamut_min}.

Here, we make two important remarks about affine gamut mapping: 1) Gamut mapping and colour correction combined can be represented by the single affine transform: $3 \times 3$ matrix $TM$ and offset $\mb{o}$; 2) It follows that the rank-based method presented in the next section will actually solve for $TM$. The offset term can be incorporated directly in $f()$ (since an offset does not change ranking).



\subsection{Rank-based estimation for colour correction}
Let us denote the $k^{\text{th}}$ row of $M$ as $M_k$, let us assume that given two colour corrected RAWs, $M_k\mb{\rho}^a$ and $M_k\mb{\rho}^b$ that the rank order is the same as for the corresponding rendered RGBs:
\begin{equation}
{P}^a_k>{P}^b_k\;\Rightarrow\;M_k\mb{\rho}^a>M_k\mb{\rho}^b \Rightarrow\; M_k(\mb{\rho}^a-\mb{\rho}^b)>0
\end{equation}
Defining the difference vector $\mb{d}^{j}=\mb{\rho}^a-\mb{\rho}^b$:
\begin{equation}
M_k\mb{d}^{j}>0
\label{eq:cc_3}
\end{equation}
where it is understood the superscript $^j$ denotes the difference vector from the $j^\text{th}$ of $\binom{n}{2}$ pairs of $n$ image pixel values. Suppose that we have a vector $M_k$ where Eqn.~\ref{eq:cc_3} holds, then the inequality cannot be true for $-M_k$. That is Eqn.~\ref{eq:cc_3} defines a half plane constraint~\cite{finlayson2016rank,ComputationalGeometry}. The vector $\mb{d}^{j}$ is perpendicular to the half-plane: any $M_k$ less than 90 degrees to $\mb{d}^{j}$ is a possible solution. Given multiple difference vectors then we have multiple half-plane constraints which taken together delimit a region in 3-space where $M_k$ must lie. Denoting the half-plane as ${\cal H}(\mb{d}^j)$, $M_k$ must satisfy:
\begin{equation}
M_k\in \bigcap_j {\cal H}(\mb{d}^j)
\label{eq:cc_4}
\end{equation}

\begin{figure}[htb]
\begin{center}
   \includegraphics[width=0.9\linewidth]{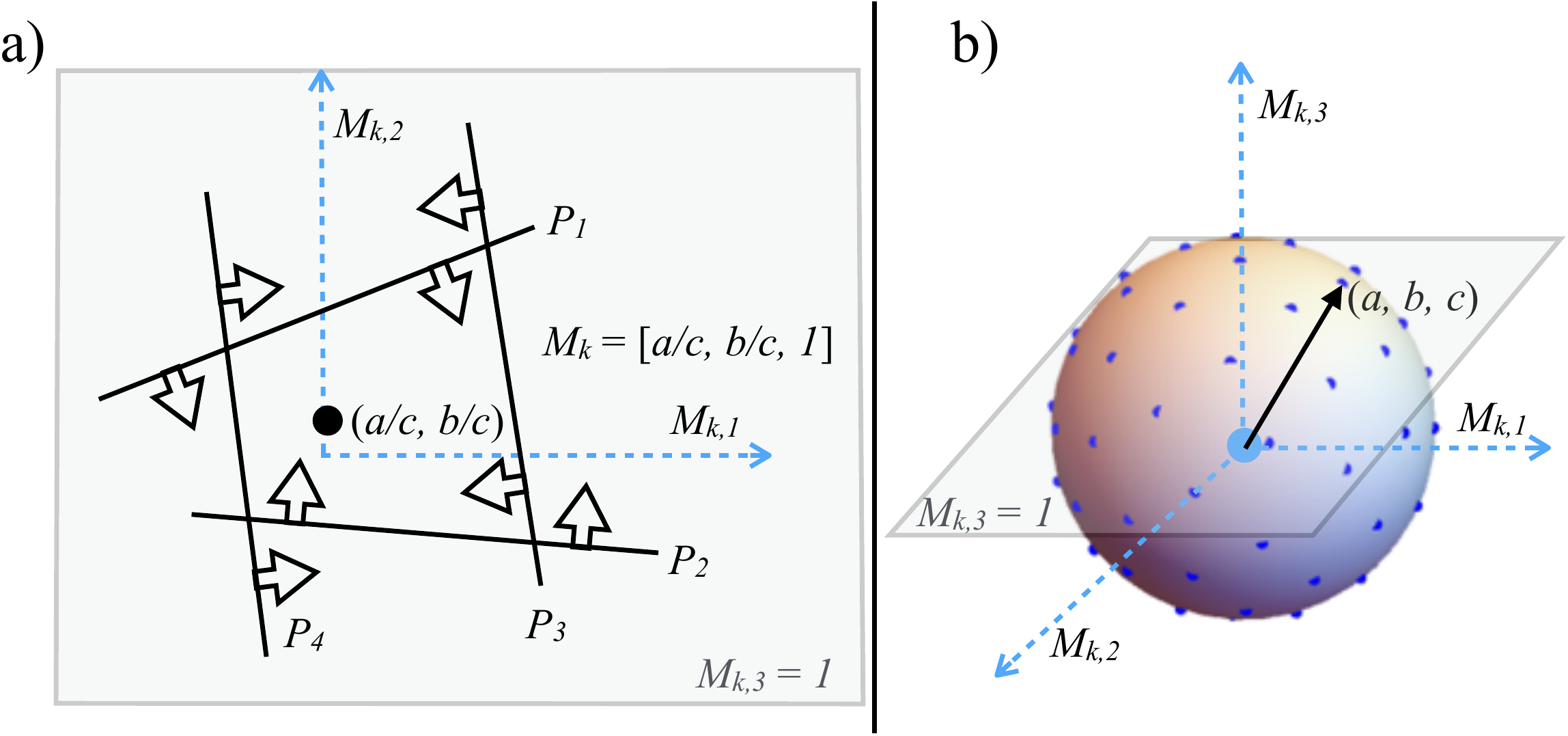}
\end{center}
\caption{a) The region where 4 half-plane constraints intersect delimit the region where $[M_{k,1},M_{k,2}]$ must lie where the black point is a feasible solution. b) On an unit sphere, each vector represented by the origin and a blue surface point is a probe for a possible solution (\eg the black arrow). All 3D points and constraints are projected to a 2D plane $M_{k,3} = 1$.}
\label{fig:half_plane}
\end{figure}

Let us visualise the computation of $M_k$ using ranking. Without loss of generality let us assume that $M_{k,3}=1$. We rewrite Eqn.~\ref{eq:cc_3} as
\begin{equation}
M_{k,1}d_1^{j}+M_{k,2}d_2^{j}+d_3^{j}>0
\label{eq:cc_5}
\end{equation}
If $[a\;b\;c]$ is a solution to Eqn.~\ref{eq:cc_4}, then $[a/c\;b/c\;c/c]$ for Eqn.~\ref{eq:cc_5} is also true since $M_{k,1}=a/c$ and $M_{k,2}=b/c$. Solutions for $[M_{k,1},M_{k,2}]$ lie on one side of the line, \ie the 3D half-space constraints maps directly to a 2D half-plane constraint. Or, if we consider the whole set of collations, the cone in 3D, defined by Eqn.~\ref{eq:cc_4}, maps to a 2D convex region~\cite{FINLAYSON.PAMI.96}. Denoting half-planes as ${\cal P}(\mb{d}^j)$ we, equivalently, solve for
\begin{equation}
 [M_{k,1},M_{k,2}]\in \bigcap_j {\cal P}(\mb{d}^j)
\label{eq:cc_6}
\end{equation}
The intersection problem of Eqn.~\ref{eq:cc_6} is easily visualised. In Fig.~\ref{fig:half_plane}a we show the intersection of 4 half plane constraints and indicate the solution set where $M_k$ must lie. 

We solve for $M$ one sensor channel at a time. Empirically, we have to be careful not to generate too many half planes. In our experiment, we generate half-planes from all pairs of up to 50 randomly selected unique RAWs, generating $2450$ half-planes. Due to noise or small deviations in real camera data, it is likely that no common intersection can be found that satisfy every half-planes constraint. To solve this problem, we generate 100,000 unit length vectors that are uniformly distributed on the surface of the unit sphere~\cite{SphericalSampling}, which is visualised in Fig.~\ref{fig:half_plane}b. With respect to this sampling, the furthest distance between any point and its nearest neighbour is less than 1.15 degrees. So, the orientation of the rows of $M$ are found to this accuracy. For each point on the sphere (\ie a possible row of $M_k$), we count how many half-space constraints are satisfied. The point on the unit sphere that satisfies most half-plane constraints -- or the median of multiple points if there is a tie -- defines $M_k$.
To make our approach robust, we find randomly select 50 colours 25 times and for each trial find the best $M_k$. Overall, we find the $M$ that places all the corresponding raw and rendered image RGBs in the most similar rank order. That is, if we plot the mapped raw red responses, for example, against the rendered red JPEG corresponding values then the graph should be a monotonically increasing function. How well a monotonically increasing function fits our data can be used to judge the efficacy of each $M$.

Ranking can only estimate $M$ up to an unknown scaling of its rows. Suppose for a rendered achromatic RGB $\mb{P}^A=[0.5\;0.5\;0.5]^\intercal$ and the corresponding raw $\mb{\rho}^A=[a\;b\;c]^\intercal$, we apply: $0.5\textbf{diag}(M\mb{\rho}^A)^{-1}M\mb{\rho}^A=[0.5\;0.5\;0.5]^\intercal$ where $\textbf{diag}()$ places a vector along the diagonal of a diagonal matrix. After this step, $M\leftarrow 0.5\textbf{diag}(M\mb{\rho}^A)^{-1}M$ maps achromatic colours correctly.
Because $M\leftarrow DM$ (where in the example, $D=0.5\textbf{diag}(M\mb{\rho}^A)^{-1}$) we might also solve for $D$ in a least-squares sense by including all colours indexed $^i$ close to the achromatic axis: $\min_D \sum_i
||DM\underline{\rho}^i-\underline{P}^i||$ (our experiment does not include this additional step).

\subsection{Rank-preserving optimization of tone curves}
We now solve for the optimal per-channel tone curves which map colour corrected RAWs to corresponding rendered RGBs. Let us denote the $i^\text{th}$ colour corrected RAW and rendered RGB pixel pairs for the $k^\text{th}$ channel as $(\rho_{k,i},P_{k,i})$. Then, the $k^\text{th}$-channel rank-preserving tone curve $f_k()$ is optimised as a $7^\text{th}$ order monotonic and smooth polynomial function as follows:
\begin{equation}
    \min_{f_k()}\Sigma_i ||f_k (\rho_{k,i})-P_{k,i}||^2 + \lambda \int_{t}||f_k''(t)||^2 dt \;\;\text{s.t.} \; f_k'()\ge0
\label{eq:im_3}.
\end{equation}
where the first term is for data fitness, the second term is for curve smoothness and $\lambda$ is a small weight (\eg$10^{-5}$). This polynomial fitting is solved by Quadratic Programming~\cite{Gill81}. 
In this paper, we further denote the combination of all 3-channel mappings $f_{1-3}()$ as $f()$.

\subsection{Gamut correction step}
As argued previously, we propose that $f(M\underline{\rho})$ has the expressive power to implement colour correction, tone correction and gamut mapping (to the first order in a Taylor expansion). However, we wish to add a further gamut mapping step for the higher order terms. But, since our hypothesis is that much of the gamut mapping will have been accounted for we are going to adopt a simple small parameter solution. Further, this additional correction is going to be carried out at the end of the process, we adopt Eqn.~\ref{eq:base_model}b.
Specifically, we find a $5\times5\times5$ LUT by using lattice regression~\cite{Lattice} that minimises $\min_{LUT()}\Sigma_i ||\text{LUT}(f(M\mb{\rho}_i))-\mb{P}_i||^2$.

\subsection{Rank-based recovery of raw}
Suppose we wish to map rendered RGBs to RAWs. Using the method presented in Section 3.2, $M$ has already been solved in the RAW-to-JPEG forward estimation phrase. Now, in a least-squares optimal way, we use the same polynomial fitting method (Eqn.~\ref{eq:im_3}) to find $f^{-1}$ by optimising $\min_{f^{-1}()} \Sigma_i ||f^{-1}(\mb{P}_i)-M\mb{\rho}_i||$. Finally, we solve for the backward $\text{LUT}()$ by optimising ${\min_{LUT()} \Sigma_i ||\text{LUT}(M^{-1}f^{-1}(\mb{P}_i))-\mb{\rho}_i ||}$ where the LUT is fitted by a $5\times5\times5$ lattice regression~\cite{Lattice}.

\subsection{Parameter counting}

Assuming we solve for 3 independent tone curves then our method requires 9 (for $M$) + $8 \times 3$ (for $f()$) + $125\times 3$ (for $\Gamma()$) = 408 parameters which is significantly less (even an order of magnitude less) than \cite{chak,BrownPami12,BrownECCVLattice12}.

\section{Evaluation}

Our evaluation is based on the most challenging dataset that we have encountered: ~\cite{chak} which contains the RAW/JPEG intensity pairs of 140 colour checker patches viewed under multiple viewing conditions. Specifically, the colour chart is captured by 8 cameras (3 of which are JPEG-only) and under 16 illuminants across many different exposures. 

Below, we carried out the same experiment described in \cite{chak}. We are interested in validating whether our method, with much reduced number of parameters can produce, similar or even better results compared with \cite{chak}. We evaluate both RAW-to-JPEG and JPEG-to-RAW.
The dataset~\cite{chak} captures a sort of ``worst-case'' viewing conditios. Normally, when we capture a picture there is a single prevailing illuminant colour. In the dataset of Chakrabarti \etal, all camera processing parameters are turned off and then the same reflectances are viewed under multiple coloured lights. As Forsyth observed~\cite{forsyth1990novel}, the reddest red camera response cannot be observed under a blue light. And, then he exploited this observation to solve for the colour of the light. Yet, in this dataset the reddest red, the greenest green and the bluest blue can all appear simultaneously in the same image. Practically, we believe the need to divine a pipeline for the all lights all surfaces case means the prior art pipelines are probably more complex than they need to be. 

As described in \cite{chak}, for each camera, we estimate the parameters of a calibration model using different subsets of available RAW-JPEG pairs. For each subset and a selected camera, the root mean-squared error (RMSE)  between the prediction and ground truth is validated by using all available RAW-JPEG pairs. Table~\ref{table:result}a shows the RAW-to-JPEG mapping error (where pixel intensities are coded as integers in the interval $[0,255]$. In the table, {\it Prob} denotes the Chakrabarti method (with several thousands parameters) and {\it RB} the rank-based method with 408 parameters.
We found that our forward errors are close to the results of \cite{chak}, especially for the condition of less than 3 illuminants which are more likely to occur in the real world. Evidently, for the many illuminant case the prior art has a small advantage. Remembering that JPEGs are coded as integers in [0,255] the RMSE is typically 1 or less (for {\it RB} compared to {\it Prob}). Practically, when the ``fits'' are viewed visually (by looking at images) it has hard to see the difference between the two methods.


For computer vision, we are more interested in the performance of JPEG-to-RAW mapping which is shown in Table~\ref{table:result}b and Table~\ref{table:result}c. In \cite{chak}, a probabilistic framework for mapping rendered RGB to raw was presented. Here we take their mean estimates as the most likely RAW predictions. We found that our method generally reduces the errors of \cite{chak} by $\sim 34\%$. Our supplementary material also includes the additional experiment results compared with ``\cite{ChakBMVC} + our LUT'' for interested readers.

The reader might be interested why our simple method seems to work so well going from rendered to raw (better than \cite{chak}) but not quite as well as the prior art in the forward direction (albeit visually almost indistinguishable). Our hypothesis here is that the LUT in the forward direction is applied post the tone curve. This curve (at least for dark values) has a very high slope and, consequently, the coarsely quantised $5\times5\times5$ LUT cannot capture gamut mapping well. Yet, in the reverse direction (JPEG to RAW) the LUT is applied in linear raw where a course uniform quantisation is more justified.

\begin{table}
\small
\centering
\begin{tabular}{lcccccccc}
\toprule
\textbf{a) RAW-to-JPEG}& \multicolumn{2}{c}{Uniform 8K}  & \multicolumn{2}{c}{10 Exp. 1 Illu.}  & \multicolumn{2}{c}{10 Exp. 2 Illu.}  & \multicolumn{2}{c}{4 Exp. 4 Illu.} \\ 
\cmidrule{2-9}
Camera & Prob & RB & Prob & RB & Prob & RB & Prob & RB\\
\cmidrule(r){1-1} \cmidrule(r){2-3} \cmidrule(r){4-5}\cmidrule(r){6-7}\cmidrule(r){8-9}
Canon\_EOS\_40D & 1.84 & 2.56 & 9.79 & 10.10 & 7.53 & 4.13 & 4.06 & 5.87 \\
Canon\_G9 & 2.17 & 3.70 & 6.51 & 6.20 & 3.41 & 5.48 & 3.09 & 4.79 \\
Canon\_PowerShot\_S90 & 2.44 & 3.24 & 4.88 & 4.52 & 3.58 & 4.34 & 3.40 & 4.04 \\
Nikon\_D7000 & 1.72 & 4.03 & 8.05 & 10.03 & 3.32 & 5.39 & 26.06 & 6.48 \\
Panasonic\_DMC-LX3 & 1.65 & 3.65 & 7.33 & 8.70 & 5.25 & 4.56 & 3.05 & 7.98 \\
\midrule
& \multicolumn{2}{c}{} & \multicolumn{2}{c}{8 Exp. 4 Illu.}  & \multicolumn{2}{c}{4 Exp. 6 Illu.}  & \multicolumn{2}{c}{8 Exp. 6 Illu.}\\
\cmidrule{4-9}
Camera & & & Prob & RB & Prob & RB & Prob & RB\\
\cmidrule(r){1-1} \cmidrule(r){4-5}\cmidrule(r){6-7}\cmidrule(r){8-9}
Canon\_EOS\_40D &  &  & 2.91 & 4.13 & 3.60 & 4.11 & 2.25 & 3.61 \\
Canon\_G9 &  &  & 2.79 & 5.48 & 3.12 & 4.74 & 2.77 & 4.67 \\
Canon\_PowerShot\_S90 &  &  & 2.95 & 4.34 & 3.27 & 3.70 & 2.75 & 3.93 \\
Nikon\_D7000 &  &  & 2.41 & 5.39 & 2.77 & 5.04 & 1.92 & 4.95 \\
Panasonic\_DMC-LX3 &  &  & 2.77 & 4.56 & 2.94 & 4.26 & 2.33 & 4.14 \\
\toprule
\textbf{b) JPEG-to-RAW} & \multicolumn{2}{c}{Uniform 8K}  & \multicolumn{2}{c}{10 Exp. 1 Illu.}  & \multicolumn{2}{c}{10 Exp. 2 Illu.}  & \multicolumn{2}{c}{4 Exp. 4 Illu.} \\ 
\cmidrule{2-9}
Camera & Prob & RB & Prob & RB & Prob & RB & Prob & RB\\
\cmidrule(r){1-1} \cmidrule(r){2-3} \cmidrule(r){4-5}\cmidrule(r){6-7}\cmidrule(r){8-9}
Canon\_EOS\_40D & 0.079 & 0.060 & 0.085 & 0.072 & 0.080 & 0.064 & 0.075 & 0.072 \\
Canon\_PowerShot\_G9 & 0.126 & 0.075 & 0.143 & 0.104 & 0.120 & 0.079 & 0.120 & 0.082 \\
Canon\_PowerShot\_S90 & 0.065 & 0.052 & 0.073 & 0.058 & 0.069 & 0.074 & 0.066 & 0.057 \\
Nikon\_D7000 & 0.143 & 0.090 & 0.543 & 0.123 & 0.140 & 0.098 & 0.229 & 0.108 \\
Panasonic\_DMC-LX3 & 0.082 & 0.058 & 0.090 & 0.072 & 0.082 & 0.063 & 0.073 & 0.071 \\
\midrule
& \multicolumn{2}{c}{} & \multicolumn{2}{c}{8 Exp. 4 Illu.}  & \multicolumn{2}{c}{4 Exp. 6 Illu.}  & \multicolumn{2}{c}{8 Exp. 6 Illu.}\\
\cmidrule{4-9}
Camera & & & Prob & RB & Prob & RB & Prob & RB\\
\cmidrule(r){1-1} \cmidrule(r){4-5}\cmidrule(r){6-7}\cmidrule(r){8-9}
Canon\_EOS\_40D &  &  & 0.071 & 0.064 & 0.077 & 0.065 & 0.069 & 0.063 \\
Canon\_PowerShot\_G9 &  &  & 0.121 & 0.079 & 0.126 & 0.076 & 0.126 & 0.080 \\
Canon\_PowerShot\_S90 &  &  & 0.069 & 0.074 & 0.063 & 0.059 & 0.066 & 0.058 \\
Nikon\_D7000 &  &  & 0.144 & 0.098 & 0.147 & 0.094 & 0.143 & 0.101 \\
Panasonic\_DMC-LX3 &  &  & 0.077 & 0.063 & 0.074 & 0.060 & 0.077 & 0.064 \\
\bottomrule
\multicolumn{5}{l}{\textbf{c) Uniform 8K (JPEG-Only Camera Test)} }  & \multicolumn{2}{c}{RAW-to-JPEG}  & \multicolumn{2}{c}{JPEG-to-RAW}  \\ 
\cmidrule{6-9}
Camera  &\multicolumn{4}{l}{Raw Proxy} & Prob & RB & Prob & RB\\
\cmidrule(r){1-1} \cmidrule(r){2-5} \cmidrule(r){6-7}\cmidrule(r){8-9}
FUJIFILM\_J10 & \multicolumn{4}{l}{Panasonic\_DMC-LX3} & 10.43 & 11.51 & 0.279 & 0.077\\
Galaxy\_S\_III &\multicolumn{4}{l}{Nikon\_D7000} & 11.34 & 13.13 & 0.114 & 0.074 \\
Panasonic\_DMC\_LZ8 & \multicolumn{4}{l}{Canon\_PowerShot\_G9} & 8.85 & 12.23 & 0.146 & 0.085 \\
\bottomrule
\end{tabular}

\caption{RMSE between ground truth and prediction for bidirectional RAW and JPEG conversions: Prob denotes \cite{chak} and RB is our rank-based method. ``Exp.'' and ``Illu.'' are respectively short for ``Exposure'' and ``Illuminant''. ``Raw Proxy'' is the camera used to capture raw for the camera which does not support raw image capturing.}
\label{table:result}
\end{table}

\section{Calibration with small numbers of parameters}

\begin{figure}[htb]
\begin{center}
   \includegraphics[width=\linewidth]{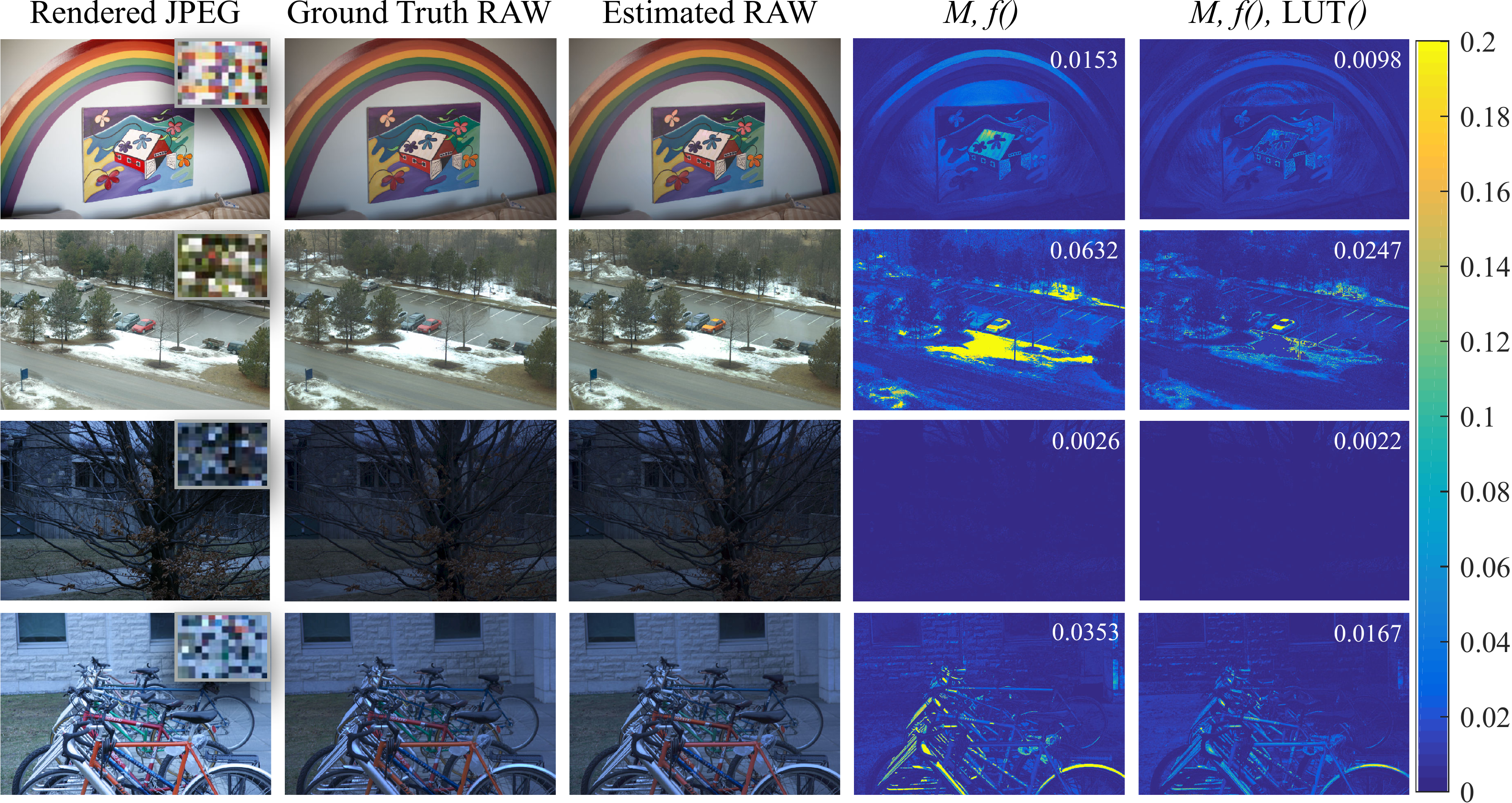}
\end{center}
\caption{Visualisation of one-shot radiometric calibration through a simulated 140-patch colour checker, shown at the top-right corner of each Rendered JPEG image. The error maps in the 4$^\text{th}$ and 5$^\text{th}$ columns respectively visualise the per pixel RMSE for our rank-based method with \& without the gamut mapping LUT. The RMSE of each whole image is shown at the top-right corner of each error map. All raw images are shown with a 0.5 gamma.}
\label{fig:one-shot}
\vspace{-20pt}
\end{figure}

We wished to visually validate our claim that we can calibrate with few parameters. We took 4 RAW+JPEG pairs (for different cameras) from~\cite{ChakBMVC}. We then uniformly selected 140 corresponding pixels from the RAW and JPEG. We solved for all parameters in our rank-based method. We then applied our model to the rest of the image. The result of this experiment for 4 images (JPEG-to-RAW) is shown in Fig.~\ref{fig:one-shot}.

\section{Conclusion}

In this paper we have shown how the rank order of image responses is a powerful tool for solving for the individual steps in a camera processing pipeline (colour correction, gamut and tone mapping). A simple ranking argument, relating colour corrected RAWs to corresponding rendered RGBs suffices to solve for the colour correction matrix. Then, the rank-preserving tone map is found and, finally, a simple gamut correction step is derived. Compared with the prior art, our rank-based method requires the fewest assumptions and delivers state-of-the-art radiometric calibration results. 

\bibliography{cvpr}
\end{document}